\documentclass{article}

\usepackage{arxiv}

\usepackage{algorithm}
\usepackage{algpseudocode}
\usepackage{amsfonts}
\usepackage{amsmath}
\usepackage{booktabs}
\usepackage{color}
\usepackage{graphicx}
\usepackage{makecell}
\usepackage{tabularx}
\usepackage{url}

\newcommand{\figwidth}[4]{
    \begin{figure}[tb]\centering
    \includegraphics[width=#4]{./fig/#1}
    \caption{#2}
    \label{#3}\end{figure}
    {}}

\usepackage{amsmath,amsfonts,bm}

\def\eqref#1{equation~\ref{#1}}

\def\1{\bm{1}}

\def\vx{{\bm{x}}}
\def\vy{{\bm{y}}}

\def\mX{{\bm{X}}}

\DeclareMathAlphabet{\mathsfit}{\encodingdefault}{\sfdefault}{m}{sl}
\SetMathAlphabet{\mathsfit}{bold}{\encodingdefault}{\sfdefault}{bx}{n}

\title{Flexible Bivariate Beta Mixture Model: A Probabilistic Approach for Clustering Complex Data Structures}

\author{
 Yung-Peng Hsu, Hung-Hsuan Chen\thanks{Corresponding author
 } \\
 National Central University \\
 \texttt{yungpeng1998@gmail.com, hhchen1105@acm.org}
 }

\date{}

\begin{document}

\maketitle

\begin{abstract}

Clustering is essential in data analysis and machine learning, but traditional algorithms like $k$-means and Gaussian Mixture Models (GMM) often fail with nonconvex clusters. To address the challenge, we introduce the Flexible Bivariate Beta Mixture Model (FBBMM), which utilizes the flexibility of the bivariate beta distribution to handle diverse and irregular cluster shapes. Using the Expectation Maximization (EM) algorithm and Sequential Least Squares Programming (SLSQP) optimizer for parameter estimation, we validate FBBMM on synthetic and real-world datasets, demonstrating its superior performance in clustering complex data structures, offering a robust solution for big data analytics across various domains. We release the experimental code at \url{https://github.com/yung-peng/MBMM-and-FBBMM}.

\end{abstract}

\section{Introduction}

Clustering is a fundamental task in data analysis and machine learning that aims to group data points into clusters such that the points in the same cluster are more similar than those in other clusters. This unsupervised learning method is widely used in various applications, including image analysis, information retrieval, text analysis, bioinformatics, and many more~\cite{wang2023detecting,lien2019visited,johnson1967hierarchical,yang2010image}. Clustering helps uncover the underlying structure of the data, facilitates data summarization, and sometimes serves as a preprocessing step for other algorithms~\cite{lien2019visited}.

Despite its widespread use, one of the primary challenges many traditional clustering algorithms face is that they often assume that the data points form clusters with convex shapes. For example, centroid-based algorithms like $k$-means and distribution-based models like Gaussian Mixture Models (GMM) typically produce clusters that are hyperspherical or ellipsoidal~\cite{hsu2024multivariate}. Although this assumption simplifies the clustering process, it restricts the flexibility of these models to handle complex data distributions that do not conform to convex shapes.

This convexity constraint can lead to suboptimal clustering results, especially when the data inherently possesses nonconvex structures. Examples include data points that form concentric circles, crescent shapes, or other intricate patterns. Traditional methods may fail to correctly group these points, leading to less optimal clustering results and loss of valuable structural information.

We propose the Flexible Bivariate Beta Mixture Model (FBBMM) to address these limitations. Unlike conventional models, FBBMM leverages the flexibility of the bivariate beta distribution, which can accommodate a wide range of shapes, including convex, concave, and other irregular forms. This adaptability is crucial for accurately capturing the proper structure of complex datasets.

The FBBMM offers several advantages over traditional clustering algorithms. First, versatile cluster shapes: FBBMM can model clusters with various shapes using the bivariate beta distribution, providing a better fit for nonconvex data structures. Second, soft clustering: like GMM, FBBMM assigns a probability to each data point to belong to different clusters, offering a better and more flexible representation of data point memberships. Third, generative capability: FBBMM, being a generative model, can generate new data points that resemble the original data, which is helpful in data augmentation and simulation tasks.

In this paper, we detail the formulation of FBBMM, describe its probability function and the parameter estimation process using the Expectation Maximization (EM) algorithm, and demonstrate its effectiveness through experiments on synthetic and real-world datasets. The results indicate that FBBMM outperforms traditional models in handling nonconvex clusters and provides a robust framework for flexible and accurate data clustering.

The rest of the paper is organized as follows. In Section~\ref{sec:rel-work}, we review famous clustering algorithms of various types. Section~\ref{sec:method} presents the bivariate beta distribution, the FBBMM model, and the parameter learning process. Section~\ref{sec:exp} compares FBBMM with famous clustering algorithms using both synthetic and open datasets. Finally, we conclude our work and discuss the limitations of FBBMM and future work in Section~\ref{sec:disc}.
\section{Related Work} \label{sec:rel-work}

Clustering algorithms can be categorized into four types: centroid-based, density-based, hierarchical, and distribution-based methods. Each has its strengths and limitations, as discussed below, followed by a comparison with our proposed Flexible Bivariate Beta Mixture Model (FBBMM).

Centroid-based methods like $k$-means~\cite{macqueen1967some} are computationally efficient but assume convex clusters, making them unsuitable for nonconvex data. Density-based methods like DBSCAN~\cite{ester1996density} identify clusters of arbitrary shapes and are robust to noise but depend heavily on hyperparameter tuning. Hierarchical methods, such as agglomerative clustering~\cite{day1984efficient}, build a tree-like structure and do not require pre-specifying cluster numbers but are computationally expensive and struggle with large datasets. Distribution-based models like Gaussian Mixture Models (GMM)~\cite{reynolds2009gaussian} handle soft clustering but are limited to elliptical cluster shapes. MBMM~\cite{hsu2024multivariate} addresses this by assuming multivariate beta distributions, allowing nonconvex clusters but restricting correlations to be positive.

FBBMM overcomes these limitations by employing the flexible bivariate beta distribution, enabling it to model both convex and nonconvex clusters and handle positive and negative correlations. It supports soft clustering and is generative, capable of producing new data points for tasks like data augmentation. Although FBBMM handles bivariate data, this limitation can be mitigated using dimension reduction techniques such as PCA or autoencoders.

\begin{table*}[tb]
\centering
\caption{Comparison of Clustering Algorithms}
\label{tab:method-cmp}
\resizebox{\columnwidth}{!}{
\begin{tabular}{@{}cccccc@{}}
\toprule
\textbf{Algorithm} & \textbf{Type} & \textbf{Shape} & \textbf{Assignment} & \textbf{Noise Robustness} & \textbf{Generative} \\ \midrule
$k$-means & Centroid-based & Convex & Hard & Low & No \\
DBSCAN & Density-based & Arbitrary & Hard & High & No \\
Agglomerative & Hierarchical & Arbitrary & Hard & Medium & No \\
GMM & Distribution-based & Convex & Soft & Low & Yes \\
MBMM & Distribution-based & Flexible & Soft & Medium & Yes \\
FBBMM & Distribution-based & Flexible & Soft & Medium & Yes \\ \bottomrule
\end{tabular}
}
\end{table*}

As shown in Table~\ref{tab:method-cmp}, FBBMM's flexibility in cluster shapes and ability to handle positive and negative correlations make it a more versatile and effective clustering method compared to traditional approaches.
\section{Flexible Bivariate Beta Mixture Model} \label{sec:method}

The Flexible Bivariate Beta Mixture Model (FBBMM) leverages the flexibility of the bivariate beta distribution to model clusters with a variety of shapes, addressing the limitations of traditional clustering methods, which often assume convex cluster shapes. In this section, we describe the FBBMM in detail, including the PDF of the flexible bivariate beta distribution, the FBBMM density function, and the parameter learning process.

\begin{table*}[tb]
\centering
\caption{Definition of Variables in FBBMM}
\begin{tabularx}{\columnwidth}{@{}cX@{}}
\toprule
\textbf{Variable} & \textbf{Definition} \\ \midrule
$N$ & Number of data points \\
$C$ & Number of clusters \\
$\vx_n$ & A data point (indexed by $n$), $\bm{x}_n = [x_{n,1}, x_{n,2}]$\\
$z_n$ & A latent variable indicating the cluster membership of $x_n$, $z_n \in \{1, 2, \ldots, C\}$\\
$\bm{\pi}$ & The probabilities of a data point belongs to the cluster $1, 2, \ldots, C$, $\bm{\pi} = [\pi_1, \ldots, \pi_C]$ \\
$\alpha_j^c$ & The $j$th parameter of the bivariate beta distribution for the cluster $c$, $j \in \{1,\ldots, 4\}, c\in\{1,\ldots,C\}$ \\
\bottomrule
\end{tabularx}
\label{tab:variables}
\end{table*}

\subsection{Bivariate Beta Distribution}

The definition of the beta distribution is unique. However, the beta distribution is only defined on a univariate variable
within the interval $[0, 1]$ or $(0, 1)$. When the number of variates is greater than one, the definition of the multivariate beta distribution is ambiguous~\cite{kotz2019continuous,hsu2024multivariate}. Eventually, we use the flexible bivariate beta distribution based on the definition provided by~\cite{olkin2015constructions} because this definition is one of the few that allows for a positive or negative correlation between covariates, making the cluster shapes more flexible.

Our bivariate beta distribution is defined based on Dirichlet distribution. Let $(U_1, U_2, U_3, U_4)$ be a set of random variables sampled from Dirichlet distributions with parameters $\bm{\alpha} = \{\alpha_1, \alpha_2, \alpha_3, \alpha_4\}$. The PDF is given by: 

\begin{equation}
f(u_1, u_2, u_3, u_4) = \frac{u_1^{\alpha_1 - 1} u_2^{\alpha_2 - 1} u_3^{\alpha_3 - 1} u_4^{\alpha_4 - 1}}{B(\bm{\alpha})},
\end{equation}
where $\alpha_{ij} \geq 0$ and $B(\bm{\alpha})$ is the normalization term, as defined below.

\begin{equation}
B(\bm{\alpha}) = \frac{\prod_{i} \Gamma(\alpha_i)}{\Gamma(\sum_i \alpha_i)}.
\end{equation}

The support of the Dirichlet distribution $u_j$s must follow the following two conditions: $0 \leq u_j \leq 1$ and $u_1 + u_2 + u_3 + u_4 = 1$. By replacing $u_4$ in the above formula with $1 - u_1 - u_2 - u_3$, we get a PDF involving three random variables:

\begin{equation}
f(u_1, u_2, u_3) = 
\frac{u_1^{\alpha_1 - 1} u_2^{\alpha_2 - 1} u_3^{\alpha_3 - 1} (1 - u_1 - u_2 - u_3)^{\alpha_4 - 1}}{B(\bm{\alpha})}.
\end{equation}

Next, we define two random variables $X$ and $Y$:

\begin{equation}
X = U_1 + U_2, \quad Y = U_1 + U_3.
\end{equation}

The marginal distribution of the Dirichlet distribution is defined as a beta distribution. Thus, the PDF of the bivariate beta distribution of $X$ and $Y$ can be written as a function involving only $u_1$ as follows.

\begin{equation}
\begin{aligned}
BBe(x, y|\bm{\alpha}) &= \int_{\Omega} f(u_1, u_2, u_3) du_1 \\
& = \frac{1}{B(\alpha)} \int_{\Omega} u_1^{\alpha_1 - 1} (x - u_1)^{\alpha_2 - 1} \quad (y - u_1)^{\alpha_3 - 1} (1 - x - y + u_1)^{\alpha_4 - 1} du_1,
\end{aligned}
\end{equation}
where $\Omega = \{u_{11} : \max(0, x + y - 1) < u_{11} < \min(x, y)\}$.

\begin{figure}[tbh]
\centering
\includegraphics[width=.65\textwidth]{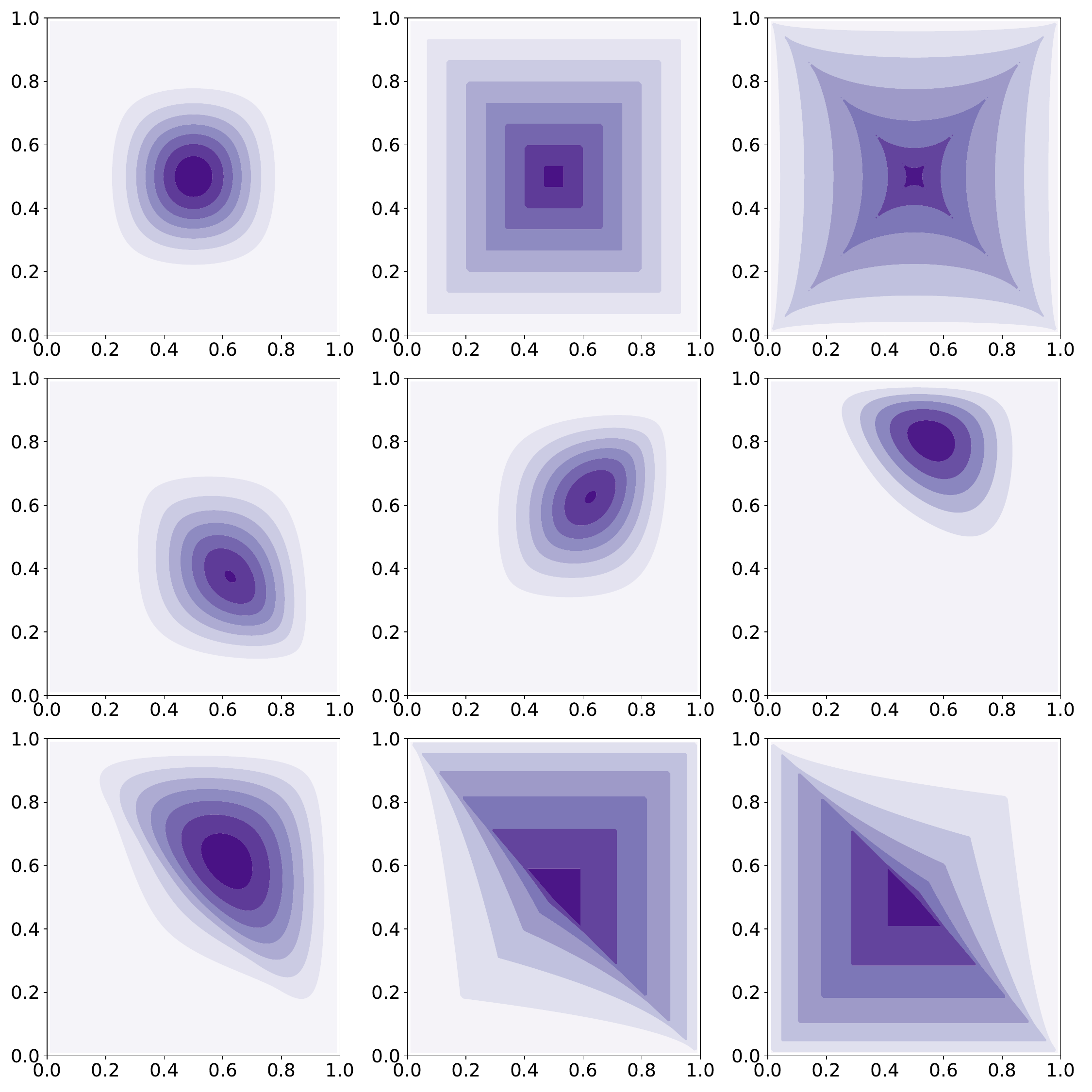}
\caption{The PDF plots of the bivariate beta distribution with different parameters. The top row: $\bm{\alpha}=(3,3,3,3)$; $\bm{\alpha}=(1,1,1,1)$; $\bm{\alpha}=(0.8,0.8,0.8,0.8)$. The middle row: $\bm{\alpha}=(2,4,2,2)$; $\bm{\alpha}=(4,2,2,2)$; $\bm{\alpha}=(4,2,4,0.5)$. The bottom row: $\bm{\alpha}=(2,2,2,0)$; $\bm{\alpha}=(1,1,1,0.5)$; $\bm{\alpha}=(0.5,1,1,1)$. The shapes could be nonconvex (e.g., upper right subfigure). The covariates could be positively correlated (e.g., the middle center subfigure) or negatively correlated (e.g., the lower left subfigure), or non-correlated (e.g., the upper middle subfigure).}
\label{fig:flex_beta_pdf2}
\end{figure}

Different parameters $\bm{\alpha}$ result in different bivariate beta distributions. The PDF mode changes according to the values of $\bm{\alpha}$. The mode is in the center if all the parameters $\alpha_j$ are equal. If the value of $\alpha_1$ becomes larger, the mode moves toward the upper right corner, i.e., the modes of the two varieties become larger. Figure~\ref{fig:flex_beta_pdf2} gives PDF examples using different $\bm{\alpha}$s.

\subsection{Generative Process and Probability Density Function of FBBMM}

\figwidth{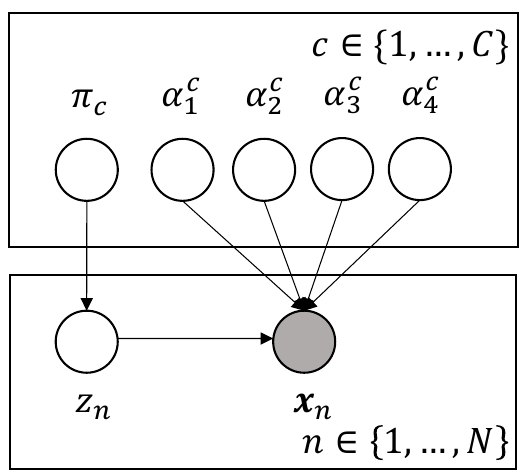}{The plate notation of the flexible bivariate beta mixture model}{fig:plate-notation}{.35\columnwidth}

We introduce the FBBMM from the perspective of a generative process. Figure~\ref{fig:plate-notation} gives the plate notations of the observed and latent variables of the FBBMM, with the notations listed in Table~\ref{tab:variables}. An observed random variable $\vx_n$ is assumed to be sampled by the following process. First, we sample a latent variable $z_n$ from a multinomial distribution with parameters $\bm{\pi}=[\pi_1, \ldots, \pi_C]$. The latent variable $z_n$ represents the cluster ID of the data point $\vx_n$. Next, $\vx_n$ is sampled from the flexible bivariate beta distribution with parameters $\alpha_{1}^{z_n}, \alpha_{2}^{z_n}, \alpha_{3}^{z_n}, \alpha_{4}^{z_n}$, the four parameters defining the bivariate beta distribution for the cluster $z_n$.

Assume that all data points $\mX = [\vx_1, \ldots, \vx_n]$ are generated from an unknown parameterized FBBMM, the PDF of FBBMM is expressed as:

\begin{equation}
p(\mX | \bm{\theta}) = \prod_{n=1}^N p(\vx_n|\bm{\theta}) = \prod_{n=1}^N \sum_{c=1}^{C} \pi_c BBe(\vx_n|\bm{\theta}_c),
\end{equation}
where $\vx_n = [x_{n,1}, x_{n,2}]$ is a 2D data point, $n \in \{1, \ldots, N\}$, $\bm{\theta}_c = \{\alpha_1^c, \alpha_2^c, \alpha_3^c, \alpha_4^c$\} are the parameters of cluster $c$, $\bm{\theta}$ includes the parameters of all clusters, and $\pi_c$ is the probability that a data point belong to the cluster $c$, thus $\sum_{c=1}^{C} \pi_c = 1$.

\subsection{Parameter Learning for FBBMM}

In practice, we only observe $\vx_1, \ldots, \vx_N$, but the other variables $\alpha_1^{1:C}, \alpha_2^{1:C}, \alpha_3^{1:C}, \alpha_4^{1:C}$, and $\pi_1, \ldots, \pi_C$ are unknown. To learn the parameters of the FBBMM, we use the Expectation Maximization (EM) algorithm. Our objective is to find the parameters $\bm{\theta}$ that maximize the likelihood function:

\begin{equation}
L(\bm{\theta}) = p(\mX|\bm{\theta}) = \prod_{n=1}^{N} p(\vx_n|\bm{\theta}) = \prod_{n=1}^{N} \sum_{c=1}^{C} \pi_c BBe(\vx_n|\bm{\theta}_c).
\end{equation}

Due to the numerical instability of multiplications when $N$ is large, we take the logarithm of the likelihood function by convention to form the log-likelihood.

\begin{equation}
\log(L(\bm{\theta})) = \sum_{n=1}^{N} \log \left( \sum_{c=1}^{C} \pi_c BBe(\vx_n|\bm{\theta}_c) \right).
\end{equation}

Assuming that we know the latent variable $z_n$, which indicates the membership of the cluster of each $x_n$, the complete log-likelihood is:

\begin{equation} \label{eq:log-prob}
\log(L(\bm{\theta})) = \sum_{n=1}^{N} \sum_{c=1}^{C} I(z_n = c) (\log \pi_c + \log BBe(\vx_n|\bm{\theta}_c)),
\end{equation}
where $I()$ is the indicator function, i.e., its output is 1 if $z_n = c$ and 0 otherwise.

In practice, since $z_n$ is unobservable, we compute $\gamma_{n,c}$, the expected probability that $x_n$ belongs to cluster $c$.

\begin{equation}
\label{eq:gamma-nc}
\gamma_{n,c} = \frac{\pi_c BBe(\vx_n|\bm{\theta}_c)}{\sum_{k=1}^{C} \pi_k BBe(\vx_n|\bm{\theta}_k)}.
\end{equation}

In the E-step of EM, we assume that all the parameters $\bm{\theta}_c$s and $\pi_c$s are correct and use them to compute $\gamma_{n,c}$ (Equation~\ref{eq:gamma-nc}). In the M-step, we update the parameters using maximum likelihood estimation. The update for $\pi_c$ is:

\begin{equation} \label{eq:pi-c}
\pi_c = \frac{1}{N} \sum_{n=1}^{N} \gamma_{n,c}.
\end{equation}

For the parameters $\bm{\theta}_c = \{\alpha_1^c, \alpha_2^c, \alpha_3^c, \alpha_4^c\}$ of each cluster $c$, we use the Sequential Least Squares Programming optimizer (SLSQP) to maximize the expected value of Equation~\ref{eq:log-prob} since there seems to be a lack of closed-form solutions.

\begin{equation} \label{eq:log-prob-exp}
E_{z_{1:N}}[\log(L(\bm{\theta}))] = \sum_{n=1}^N \sum_{c=1}^C \gamma_{n,c} (\log \pi_c + \log BBe(\vx_n|\bm{\theta}_c)).
\end{equation}

The algorithm~\ref{alg:fbb-learn} provides the pseudocode for parameter learning in FBBMM.

\begin{algorithm}[tb]
\caption{FBBMM Parameter Learning}
\label{alg:fbb-learn}
\begin{algorithmic}[1]
\State \textbf{Input:} $\mX = \{\vx_1, \ldots, \vx_N\}$: $N$ input data points, each data point $\vx_n$ is 2-dimensional; $C$: the number of clusters
\State \textbf{Output:} Final parameters $\bm{\theta}_{1:C} = \{\alpha_1^{1:C}, \alpha_2^{1:C}, \alpha_3^{1:C}, \alpha_4^{1:C}\}; \bm{\pi} = \{\pi_1, \ldots, \pi_C\}$
\State Initialize parameters $\bm{\theta}_{1:C}$ and $\bm{\pi}$
\State $\text{Old\_prob} \gets -\infty$
\For{$i = 1$ to \text{Epochs}}
    \State // E-step
    \State Compute each $\gamma_{n,c}$ by Equation~\ref{eq:gamma-nc}
    \State Compute $\text{New\_prob}$ by Equation~\ref{eq:log-prob}
    \If {$|\text{New\_prob} - \text{Old\_prob}| < \epsilon$}
        \State break
    \EndIf

    \State $\text{Old\_prob} \gets \text{New\_prob}$
    
    \State // M-step
    \State Compute $\alpha_1^{1:C}, \alpha_2^{1:C}, \alpha_3^{1:C}, \alpha_4^{1:C}$ by maximizing Equation~\ref{eq:log-prob-exp} using SLSQP
    \State Compute each $\pi_c$ using Equation~\ref{eq:pi-c}
\EndFor
\end{algorithmic}
\end{algorithm}
\section{Experiments} \label{sec:exp}

This section presents the results of experiments that compare the performance of FBBMM with baseline clustering algorithms on different datasets. The compared methods include $k$-means, MeanShift, DBSCAN, Agglomerative Clustering, GMM, and MBMM. The experiments were carried out on synthetic and real-world datasets, including a structural dataset and an image dataset.

\subsection{Experimental Setup}

We preprocess the data such that the value of each feature is normalized: let $\vx_n = [x_{n,1}, \ldots, x_{n,m}]$, each $x_{n,j}$ is normalized below.

\begin{equation}
x_{n,j} = 0.01 + \frac{(x_{n,j} - \min(x_{*,j}))(0.99 - 0.01)}{\max(x_{*,j}) - \min(x_{*,j})},
\end{equation}
where $x_{*,j} = [x_{1,j}, x_{2,j}, \ldots, x_{N,j}]$, i.e., the $j$th feature of all instances.

\subsection{Experiments on the Synthetic Datasets}

The synthetic datasets were generated using scikit-learn to test the characteristics of different clustering algorithms. These datasets consist of five different shapes. Each dataset includes 500 two-dimensional data points.

\begin{figure}[tb]
\centering
\includegraphics[width=.91\textwidth]{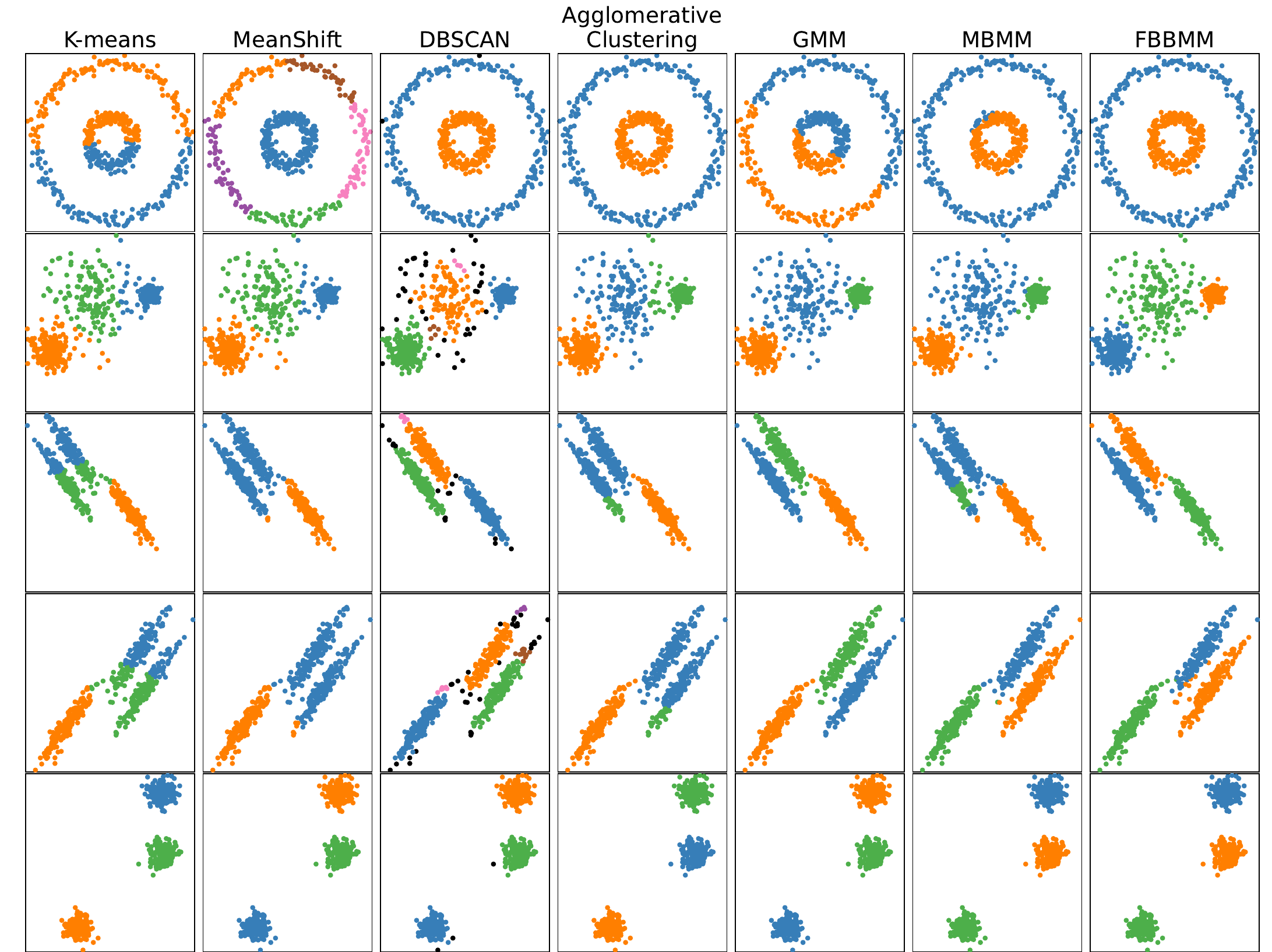}
\caption{Clustering results on synthetic datasets}
\label{fig:synthetic_results}
\end{figure}

Figure~\ref{fig:synthetic_results} compares the clustering results of $k$-means, MeanShift, DBSCAN, Agglomerative Clustering, GMM, MBMM, and FBBMM on five synthetic datasets. 

The first dataset includes concentric circles. If a point from the outer circle is selected, the most distant data point is positioned on the opposite side of the same circle. This characteristic makes the synthetic dataset highly challenging for centroid-based and distribution-based methods to group the entire outer circle into a single cluster. As shown in the first row of Figure~\ref{fig:synthetic_results}, density-based algorithms (DBSCAN) and Hierarchical clustering method (Agglomerative Clustering) and two beta distribution-based models (MBMM and FBBMM) successfully separate the two circles. 

The second dataset contains two distant 2D Gaussian distributions with small variances in each dimension, and the third distribution has a large variance, located in the middle. Thus, several data points sampled from the third distribution are mixed with the first two distributions. Since the middle cluster has a wider spread, the centroid is far from some points within the same cluster, making certain clustering algorithms, $k$-means, MeanShift, and DBSCAN, misidentify some data points in the middle cluster as other clusters; details are in the second row of Figure~\ref{fig:synthetic_results}.

The third and fourth datasets each comprise three 2D Gaussian distributions with isolated means. However, the two covariates are highly correlated: the covariates are negatively correlated for the third dataset and positively correlated for the fourth. As a result, data points are sometimes closer to those generated from other distributions. Thus, $k$-means, MeanShift, DBSCAN, and Agglomerative Clustering make errors on some data points. MBMM only handles data points whose covariates are positively correlated~\cite{hsu2024multivariate}. FBBMM and GMM are the only models that handle the two datasets well, as presented in the third and fourth rows of Figure~\ref{fig:synthetic_results}.

Finally, the last dataset includes data points from three 2D Gaussian distributions with distant means and small variances in each dimension. Therefore, a data point is close to other data points within the same Gaussian distribution but far from others. All clustering algorithms perform well in this ideal case.

Overall, our proposed FBBMM performs well on all synthetic datasets.




\subsection{Experimented the Open Datasets}

The open datasets include the wine dataset~\cite{aeberhard1994comparative} and the MNIST dataset~\cite{lecun1998gradient}. The wine dataset contains chemical analysis results of wines grown in the same region of Italy but derived from three different cultivars. There are 178 instances with 13 features. The second dataset, the MNIST dataset, comprises 70,000 grayscale images of handwritten digits (0-9), with each image having 28x28 pixels. The two datasets represent structural data and image data, respectively.

\subsection{Evaluation Metrics for Open Datasets}

We evaluate clustering results using three metrics: Clustering Accuracy (CA), Adjusted Rand Index (ARI), and Adjusted Mutual Information (AMI).

Clustering Accuracy is calculated as the number of correctly clustered data points divided by the total number of data points. Since clustering results and actual labels may not directly correspond, a mapping is performed before computing accuracy. For example, assume that we have a dataset with four data points whose labels are $[a,a,b,b]$, and a clustering algorithm produces the output with cluster IDs $[b,b,a,a]$. Despite the mismatch between the cluster IDs and the actual labels, the clustering is perfect because all points with the actual label $a$ are grouped in cluster b and vise versa. We call a set of lists the \emph{identical lists} if one list can be transformed into another list by permuting the labels. Thus, clustering accuracy is defined as the maximum accuracy among all identical lists of predicted cluster IDs~\cite{chen2023toward}.

\begin{equation} \label{eq:clusteracc}
CA\left(\vy, \bm{\hat{y}}\right) := \max_{\forall {\bm{\check{y}}} \in P(\bm{\hat{y}})} \left\{\frac{1}{N}\sum_{i=1}^N I\left(\check{y}_i = y_i\right)\right\},
\end{equation}
where $\vy = [y_1, \ldots, y_N]$ is the list of ground-truth labels, $\bm{\hat{y}} = [\hat{y}_1, \ldots, \hat{y}_N]$ is a list of predicted cluster IDs, $P(\bm{\hat{y}})$ returns a set of all identical lists for $\bm{\hat{y}}$, $I()$ is an indicator function, and $\bm{\check{y}} = [\check{y}_1, \ldots, \check{y}_N]$ is an identical list of $\bm{\hat{y}}$.

We also use the Adjusted Rand Index (ARI) and  Adjusted Mutual Information (AMI) for evaluation. ARI and AMI are biased toward different types of
clustering results: ARI prefers balanced partitions (clusters with similar sizes),
and AMI prefers unbalanced partitions~\cite{romano2016adjusting,chen2023toward}.

\subsection{Results on the Open Datasets}


Table~\ref{tab:wine_original} compares the clustering performance of FBBMM with six baseline clustering methods on the wine dataset. Since each of the six baseline methods can handle datasets with any number of features, we use the entire 13 features provided in the wine dataset. However, because FBBMM only handles datasets with bivariate variables, we use an autoencoder to reduce the original feature to two dimensions. As shown, FBBMM outperforms all baseline clustering methods.

\begin{table}[tb]
\centering
\caption{Clustering on Original Features (13 Dimensions) and FBBMM (2 Dimensions)}
\label{tab:wine_original}
\begin{tabular}{@{}lccc@{}}
\toprule
\textbf{Method} & \textbf{CA} & \textbf{ARI} & \textbf{AMI} \\ \midrule
k-means & 0.702 & 0.371 & 0.423 \\
MeanShift & 0.697 & 0.469 & 0.483 \\
DBSCAN & 0.506 & 0.297 & 0.380 \\
Agglomerative Clustering & 0.674 & 0.371 & 0.436 \\
GMM & 0.725 & 0.435 & 0.436 \\
MBMM & 0.719 & 0.391 & 0.389 \\
FBBMM (2D) & \textbf{0.983} & \textbf{0.947} & \textbf{0.927} \\ \bottomrule
\end{tabular}
\end{table}

We also project the dataset from the original 13-dimensional to 2-dimensional dataset using an autoencoder and apply baseline clustering algorithms on the 2-dimensional dataset. In doing so, we ensure a consistent evaluation environment that isolates the effects of the dimensionality reduction, enabling us to accurately assess the strengths and weaknesses of each method in this specific context.

Table~\ref{tab:wine_reduced} compares clustering performance on the wine dataset after dimension reduction. Probably because of autoencoder's ability in extracting key feature combinations, all baseline methods improved. FBBMM still performs the best in all three evaluation metrics, demonstrating its superiority in clustering.

\begin{table}[tb]
\centering
\caption{Clustering on Reduced Features (2 Dimensions)}
\label{tab:wine_reduced}
\begin{tabular}{@{}lccc@{}}
\toprule
\textbf{Method} & \textbf{CA} & \textbf{ARI} & \textbf{AMI} \\ \midrule
k-means (2D) & 0.961 & 0.882 & 0.860 \\
MeanShift (2D) & 0.961 & 0.882 & 0.860 \\
DBSCAN (2D) & 0.910 & 0.846 & 0.833 \\
Agglomerative Clustering (2D) & 0.978 & 0.930 & 0.900 \\
GMM (2D) & 0.949 & 0.847 & 0.833 \\
MBMM (2D) & 0.809 & 0.526 & 0.588 \\
FBBMM (2D) & \textbf{0.983} & \textbf{0.947} & \textbf{0.927} \\ \bottomrule
\end{tabular}
\end{table}

The MNIST dataset consists of 70,000 grayscale image. Due to the high dimensionality and a convolutional neural network (CNN)'s ability to handle images, we use a CNN as the feature extractor, followed by applying an autoencoder for dimension reduction. Since we are dealing with a clustering algorithm, we need to prevent CNN from learning information from the labels in the MNIST dataset. Thus, we use fashion-MNIST~\cite{lecun1998gradient} to train a CNN. After training, we remove the last fully connected layer. Then, we pass MNIST to this trained CNN to convert an image into a $1\times 512$ dimensional vector. Subsequently, we feed this vector into an autoencoder to reduce the features to 2 dimensional.

Table~\ref{tab:mnist_17} shows the clustering performance in clustering digits 1 and 7 after the feature reduction. FBBMM, again, achieves the best performance in all metrics, demonstrating its effectiveness in handling the digit recognition task.

\begin{table}[tb]
\centering
\caption{Clustering on MNIST Digits 1 and 7}
\label{tab:mnist_17}
\begin{tabular}{@{}lccc@{}}
\toprule
\textbf{Method} & \textbf{CA} & \textbf{ARI} & \textbf{AMI} \\ \midrule
k-means (2D) & 0.971 & 0.889 & 0.813 \\
MeanShift (2D) & 0.975 & 0.903 & 0.832 \\
DBSCAN (2D) & 0.944 & 0.857 & 0.761 \\
Agglomerative Clustering (2D) & 0.973 & 0.897 & 0.823 \\
GMM (2D) & 0.970 & 0.883 & 0.806 \\
MBMM (2D) & 0.930 & 0.738 & 0.633 \\
FBBMM (2D) & \textbf{0.976} & \textbf{0.907} & \textbf{0.841} \\ \bottomrule
\end{tabular}
\end{table}




\section{Discussion and Future Work} \label{sec:disc}

This paper introduces the Flexible Bivariate Beta Mixture Model (FBBMM), a novel probabilistic clustering model leveraging the flexibility of the bivariate beta distribution. Experimental results show that FBBMM outperforms popular clustering algorithms such as $k$-means, MeanShift, DBSCAN, Gaussian Mixture Models, and MBMM, particularly on nonconvex clusters. Its ability to handle a wide range of cluster shapes and correlations makes it highly effective.

FBBMM offers several advantages. Its use of the beta distribution allows for flexible cluster shapes, capturing complex structures more accurately than traditional models. It supports soft clustering, assigning probabilities to data points for belonging to clusters, which is versatile for overlapping clusters. Additionally, FBBMM is generative, capable of producing new data resembling the original dataset, useful for tasks like data augmentation and simulation.

However, FBBMM has limitations, including higher computational complexity due to iterative parameter estimation. Future work could focus on improving efficiency through parallelization or better optimization strategies, extending FBBMM to multivariate data, and enhancing robustness to noise and outliers. Applying FBBMM in diverse domains such as bioinformatics and image analysis could further validate its versatility and impact.

\section*{Acknowledgement}
We acknowledge support from National Science and Technology Council of Taiwan under grant number 113-2221-E-008-100-MY3. We thank to National Center for High-performance Computing (NCHC) of National Applied Research Laboratories (NARLabs) in Taiwan for providing computational and storage resources.

\bibliographystyle{unsrt}  
\bibliography{ref}

\end{document}